# Multi-agent evolutionary systems for the generation of complex virtual worlds

J. Kruse and A.M. Connor*

Auckland University of Technology, Colab (D-60), Private Bag 92006, Wellesley Street, Auckland 1142, New Zealand.

**Abstract**

Modern films, games and virtual reality applications are dependent on convincing computer graphics. Highly complex models are a requirement for the successful delivery of many scenes and environments. While workflows such as rendering, compositing and animation have been streamlined to accommodate increasing demands, modelling complex models is still a laborious task. This paper introduces the computational benefits of an Interactive Genetic Algorithm (IGA) to computer graphics modelling while compensating the effects of user fatigue, a common issue with Interactive Evolutionary Computation. An intelligent agent is used in conjunction with an IGA that offers the potential to reduce the effects of user fatigue by learning from the choices made by the human designer and directing the search accordingly. This workflow accelerates the layout and distribution of basic elements to form complex models. It captures the designer's intent through interaction, and encourages playful discovery.

**Keywords:** evolutionary computation, genetic algorithms, autonomous agents, multi-agent systems, interactive design.

*Corresponding author. Email:andrew.connor@aut.ac.nz

## 1. Introduction

Convincing computer graphics models are a necessity for the creation of successful games, movies and virtual reality environments. Some natural and architectural objects of higher complexity intensify this problem as they necessitate fine detail and a large number of smaller elements which act as the building blocks of a more complex whole. Designing high-quality content is a laborious and costly task that requires substantial skill, time and resources [1, 2] and often a large number of iterations are necessary to achieve the desired results. This research addresses these issues by use of human-centric evolutionary computation combined with autonomous agents in order to determine whether this process can be facilitated by semi-autonomous approaches [3].

This paper describes an Interactive Genetic Algorithm (IGA) that is driven by user input that works in conjunction with a computational software agent that supports the user in the decision making process. By shifting the workload from the human user to the computational agent, the laborious tasks of modelling are simplified and the process is partially automated.

This paper describes the design of a hybrid intelligent system to support interactive design and then utilises procedural city design as an example to demonstrate the process, identify potential benefits, and find possible issues of hybrid intelligent systems in design contexts.

## 2. Background and related work

The problem discussed in this paper deals with both Design and Artificial Intelligence, both of which are very broad fields of research. To narrow these further, and to address only the core matters of this enquiry, Generative Design (as part of Design), as well as Genetic Algorithms and Agents are discussed in more detail in the following sections.

### 2.1. Generative Design

Generative Design, often also referred to as Procedural Design, is the area of form and shape finding by use of algorithmic help [4]. It is the overarching field in which form finding is located in. There are some significant differences between manual design aided by computer software and automated design provided by software.

Computer Aided Design (CAD) systems and other tools to create three dimensional objects with help of computers are technically based on algorithms and program code. However they require the designer to manually operate the software and provide the inputs necessary to create a shape and form. Some of the processes may be partially automated, but the designer still needs to draw the objects on screen and parametrize them. These objects are often primitives such as circles, rectangles, cubes, spheres and other two or three dimensional shapes, which when combined, form the desired complex object [5]. While this manual process lays the foundation for Generative Design, it would still be considered a potentially onerous iterative process. The goal of this research is therefore to investigate automated or semi-automated design processes driven by algorithms to reduce the manual effort required.

Generative Design is the process of writing or applying often simple and small fragments of code that show objects on screen automatically, without the necessity to have the designer create underlying shapes in the first place [6]. The creation of shapes is done by software, driven by an algorithm. The combination of many simple shapes creates larger compound structures.

These two or three dimensional structures (or objects) tend to be rather complex, given the simplicity of a few lines of code [2]. For example, to create a complex procedural structure made of hundreds or thousands geometric primitives such as cubes and spheres, only 8 lines of code are required. These structures also often use recursive elements, i.e. functions or procedures that call themselves over and over again and therefore assemble a complex object from small, identical building blocks. This is similar to some plants such as ferns or trees, which are complex structures that are made from very few number of simple individual elements [7]. In case of the fern for example, the fractal or recursive nature is visible from a large scale to a microscopic level. The same shape is found over and over again, from plant to branch, and from branch to leaf, and so forth. This is usually referred to as self-similarity [8]. In the case of a procedurally modelled city, it is possible to apply some very similar approaches. While not fully self-similar, they are still based on repetition of simple building blocks, which when used in large numbers resemble rather complex structures. An example would be windows on an office building, which are simple elements but make up most of a large structure. Another example would be streets. While most streets have similar building blocks, as a whole they present a very complex and large system of gaps between buildings.

There is little evidence in the literature that a computer generated city has been made using Generative Design driven by the user in conjunction with algorithmic help. While there have been attempts to create procedural three dimensional cities as laid out by Parish and Müller [9], these were entirely computer driven and provided no user interaction. This leads to a computer generated city as such, but does not enable the outcome to reflect the designer's intent. The city is a result of the programmer's imagination and it can therefore be argued that it is similar to manual Computer Aided Design, with the difference that the user (or programmer) does not draw objects on screen, but writes code to create them. This research follows a different approach in that the user is influences the design by choosing preferred layouts.

Parish and Müller [9] allow the user simply to run the software, which then produces a random result, which may or may not resemble the designer's vision. There have also been other studies into components of generatively designed cities, for instance street structures or building structures [10], but these approaches also did not consider interactive user input. Therefore, this research aims to fill a significant gap in the existing literature by using Generative Design driven by user interaction to create complex structures while seeking to reach the designer's original vision.

Using computers to explore the space of possible images, sculptures or other complex artistic forms such as musical compositions, has enabled researchers and artists to evolve pieces of art and led to the exploration of new ideas. These range from simple arbitrary colour blobs to working functional forms such as boat designs, architectural forms or electronic circuits. Designers are being enabled to study more solutions in less time and to find forms that are outside the conventional and expand their conceptual understanding. Evolutionary approaches have also led to new methods and principles, which can be exploited in future designs [11].

Bohnacker [4] demonstrates a variety of generated typographic and abstract graphics. Other examples include generated art using L-Systems [12], generative design using very simple autonomous agents [2] and studies in architecture [7] to name a few. Generative Design has become more common for a variety of reasons, including a vast growth of computing power.

## 2.2. Evolutionary computation

Evolutionary computation borrows ideas from Darwin's theory of evolution, which states that individuals as part of a population increase their chances of survival and reproduction by way of natural selection. This selection process allows for small variations of each individual's properties, which are then passed on to the next generation through inheritance. Darwinism in conjunction with Mendel's concept of genetics formed what is known in biology as modern evolutionary synthesis [13].

Biological evolution encapsulates the following concepts [14]:
- *DNA* (Deoxyribonucleic acid) is the molecular structure that encodes the genetic information of each cell of all living organisms. It is represented as a double helix.
- *Chromosomes* are strings of DNA.
- *Genotype* is the hereditary information encoded in the DNA.
- *Phenotype* is the observable properties as a result of the DNA.
- *Reproduction* is the creation of offspring by (usually) two parents, inheriting parts from both parents' DNA.
- *Crossover* is the process of synthesizing an offspring DNA, creating a new chromosome.
- *Mutation* is a small accidental change in the offspring DNA, potentially resulting in slight variations to a straight, non-mutated crossover. Mutation happens with a very low probability.
- *Survival of the fittest* is the concept of only the strongest properties of a DNA being sustained over many reproduction cycles. Weaker DNA properties could result in weaker offspring, which in turn has a lower chance of survival. Over many generations, this leads to the elimination of weak DNA. This is also used synonymously with the term *Evolution* in the literature [14].

Modern evolutionary synthesis serves as the foundation for the many different types of evolutionary computation. While evolutionary computation borrows ideas and the notion of biological evolution from the natural process, evolutionary computation is merely an abstraction of evolutionary synthesis to emulate soft intelligent behaviour in computer software. The concepts were applied in different ways and evolved over time, so that there are now a multitude of different algorithms, which all borrow from the underlying idea of natural evolution. Examples are Genetic Algorithms [15], Evolution Strategies [16] and Genetic Programming [17]. While these all simulate natural evolution to an extent, they differ significantly in how they apply the evolutionary principles.

Genetic Algorithms are heuristic search algorithms, used to find a solution in the space of all possible solutions. Evolution Strategies are designed to find solutions to technical optimization problems [18], and Genetic Programming generates computer programs that in turn attempt to solve the actual problem [19]. Genetic Programming therefore programs computers by finding an optimal set of rules or section of code. Evolutionary Computation is the field of research that is concerned with computation based on the concepts of natural evolution.

**Genetic Algorithms**
Genetic Algorithms (GA), being part of the heuristic optimization or search algorithms, are very popular due to their relative simplicity, and are also well researched and understood [20]. They were introduced in the 1970s by John Holland and mimic natural evolution, normally by abstracting the chromosomes into binary digits [15]. These chromosomes are passed on from one population to a new population after genetics-inspired processes of crossover and mutation. An evaluation function called fitness function is then applied to establish each chromosome's performance towards the final goal. If the chromosome performs poorly, it is likely to be dropped from the pool of future 'parents'. Otherwise, if the chromosome's fitness is high, it is more likely to be selected for reproduction. The actual reproduction process is performed by using a crossover operator, which mixes parts of two parent's chromosomes to form the new child's chromosome. Finally, a mutation operator is applied to some of the new found chromosome in order to ensure a certain variation of the child's properties. This mutation operator randomly changes the value of individual digits of the chromosome binary string. Mutation operator and crossover operators effectively represent the probability of each operation (mutation and crossover) occurring. The process of simulating natural evolution is repeatedly applied for many generations and as a result, the fittest members of a population dominate, while the less fit become extinct. The underlying mechanisms of Genetic Algorithms are very simple, yet capable of showing seemingly complex behaviour and the ability to solve difficult problem sets [18].

While simple search and optimization algorithms such as hill climbing or gradient descent might have a tendency to get stuck in local maxima or minima, Genetic Algorithms avoid this issue due to their inherent creation of diversity by mutation. Genetic Algorithms are highly effective in many cases, and given that a robust fitness function and solid parameters for crossover and mutation have been selected, tend to avoid local optima in favour of a global solution [21]. Genetic Algorithms have successfully been applied to a range of different areas such as Engineering, Arts and Computer Science. Some examples include the optimization of machinery [22], evolved particle systems [2], generative jazz music [23] or optimizing the weights of neural networks [21]. Genetic Algorithms have also been successfully applied to much simpler, but somewhat similar design problems as presented in this paper, for instance finding coloured blobs and stripes that reflect the intent of the designer in the solution space of all possible combinations of colour blobs and stripes [24]. This research seeks to extrapolate the positive results to the more complex design issue in relation to Procedural City models.

Whilst there are many contradictory studies [25-28], there is a body of evidence that suggests that Genetic Algorithms are at least as effective as other metaheuristic search algorithms [29]. Such a view is supported by Li & Kou [30] who assert that Genetic Algorithms are implicitly parallel, robust and scalable, as well as powerful in global search and optimization.

**Interactive Genetic Algorithms**
Genetic Algorithms (GA) are relatively easy to implement and can be very effective if the solution space is very large. While GA have some disadvantages such as the tendency to converge towards local maxima if not properly tuned [31, 32] and the requirement of a fitness function, which can be a challenge if soft factors such as aesthetics come into play [33], GAs have been well studied and well documented.

The GA was selected for this study based on its simplicity, but the specific implementation does not use a mathematical fitness function. Instead it integrates the user (designer/artist) into the process. The user provides the necessary fitness evaluation in the selection stage of the GA, a process known as Interactive Evolutionary Computation (IEC) [34]. The underlying idea of IEC was first introduced by Dawkins in the third chapter of his book "The Blind Watchmaker". Dawkins demonstrates the process of evolution based on Darwinian theories in a software program called "biomorphs", which uses human interaction to evaluate factors such as aesthetics, appeal or attractiveness [35]. Karl Sims [36] has taken this idea of user-computer interaction for Genetic Algorithms further and suggests that the human user does not have to understand the underlying process of creating the candidates, while still being able to produce results of high complexity. He argues that such interactive evolution enables the computer as well as the human to achieve results that neither could have produced on their own.

The term 'Interactive Evolutionary Computation' was finally formed by Takagi [33], who also evaluated IEC in context of several different fields of research as a method to integrate computational optimization and human evaluation. Beside the ability to combine both optimization and evaluation for design subjects, IEC can also offer a significant benefit over other computational design methods such as Genetic Algorithms or manual computer aided design. The user can change their evaluation during the evolutionary process and drive the resulting populations into a different direction. This could potentially lead to the discovery of previously unknown outcomes and expose features that were not expected initially. Changing the objectives in regular Genetic Algorithms would require re-coding and is neither practical nor efficient. IEC allows for alterations on the fly and as a result has been recognized as a 'novelty generator' [37].

It is important to note, that the user does not necessarily evaluate the genotype or phenotype of the candidates of the Genetic Algorithm directly, but a different representation which is easier to grasp. For example, instead of a numerical bit string (genotype) or the associated colour values (phenotype) of some elements of an image, a user might select whole images that are based on both elementary parts [34]. This is an approach chosen for this research where the user is able to select rendered images which are each based on an underlying set of parameters. These parameters are encoded into a collection of numerical strings, the DNA. This DNA is then modified (mutated) and crossed over by the Genetic Algorithm. The resulting parameter set is rendered back as a new image, which is then presented to the user for consideration in the next generation.

Whilst this is an elegant way to both capture the user's preference and also avoid forming a mathematical function for aesthetics or user preference, it poses a significant problem which is related to the nature of humans. After a number of iterations, human users tend to fatigue and get slower to select, get distracted more easily and lose concentration due to the high number of visual triggers [38]. The effects of time saving while defining a computational solution for the design problem could potentially result in a less effective and less successful overall outcome of the interactive evolutionary process due to this fatigue effect. In the following section discusses autonomous agents as a possible solution to this inherent problem of IEC.

Interactive Evolutionary Computation offers significant benefits over non-interactive approaches, as it removes the necessity to find a mathematical solution for the fitness function. The intractable nature of writing an equation for aesthetics, taste and preference is elegantly avoided. Instead, the user is employed to directly provide the selection of potential candidates, which makes this approach suitable for this research.

## 2.3. Autonomous agents

Research into *agents* or *autonomous agents* is a relatively young field, which has been studied for about the past two decades. Most publications from the early 1990s presented agent definitions that are still valid and that build the foundation of our current understanding of the field.

In essence for this research it is assumed that an agent is acting in some environment or is part thereof. It is capable of deriving inputs from its environment and act accordingly in an independent, autonomous manner. Furthermore, an agent runs over a period of time, until it finishes its task and not necessarily when the human user decides to stop it. Some agents might even act beyond the control of any human user [39].

Russell and Norvig [40] classify agents into five groups based on the agent's level of intelligence and capability:

- *Simple reflex agents* act based on their current perception and their function follows the condition-action rule, which is usually implemented as a simple if-then decision. They require a fully observable environment in order to succeed.
- *Model-based reflex agents* differ from the above mainly through their ability to handle partially observable environments. They store descriptions of the un-observable environment and act similar to the reflex agent following a condition-action rule.
- *Goal-based agents* are model-based and use a database of desirable situations for their decision making process. The agents simply choose one of the multiple possibilities that lead to the desired goal.
- *Utility-based agents* store goal-states and non-goal states, from which they choose the most desirable state. This decision is made based on a utility function, which maps the state to a measure of the utility of a particular state.
- *Learning agents* are able to operate in unknown (or non-observable) environments through learning. These agents become more knowledgeable over time, compared to their initial state.

This research specifically involves Learning Agents in order to capture the user's intent, which is unknown at the time of programming. This therefore presents a set of non-observable parameters of the agent's environment when the process of form finding is started. But through observation of the user's action and accordingly through evaluation of success and failure by comparison of the agent's prediction and the user's input, the agent will become more knowledgeable. The idea of learning as opposed to simple behaviour is that perception is not only used to trigger certain actions depending on the observed changes in the environment, but that it is used to improve future decisions by the agent system. It is not a reaction to the environmental change, but rather a reaction to the agent's own experience [40].

Many different forms of learning are being used in agent research. Some examples include Decision Trees, which is learning from observations to generate a decision hierarchy, expert systems, which extract rules from examples or Reinforcement Learning, which is learning the value of actions by getting rewards or punishment depending on previously made decisions and applying these updated learned values to future actions [41]. In general, learning can be classified into three main categories, namely supervised, unsupervised and reinforcement learning [40].

Supervised learning is based on examples, which are used as a training dataset to teach the system. This training set includes the right and wrong answers to a problem, which the system then learns as a function of inputs and outputs, or in other words as a relationship between actions and outcomes. This can be as simple as detecting whether an image contains a certain element [2] or which action to take, when a certain event occurs in the observed environment [42]. It is important to note, that supervised learning does not require a teacher to provide the actual value for the correct solutions to the agent. The solution can also be derived by the agent from looking at all possible candidates through its own perception and getting the correct solution pointed out by the teacher. The difference is that the former requires some sort of table or key-value pairs for all right and wrong solutions, whereas the latter just requires someone to point to the right ones. This means that an agent in a fully observable environment might be able to perceive the consequences of its decisions, and learn from them to make future decisions. In a partially observable environment, this is more difficult and the agent needs more comprehensive feedback from the teacher in order to make future predictions [40].

Unsupervised learning differs in that it requires detection of patterns in the observations, because the right and wrong solutions are not provided prior to the decision making process. Examples of unsupervised learning methods are statistical learning methods or some neural network implementations. Neural networks imitate the processes in the brain by using multiple simple units with inputs and outputs called *neurons* or in their simplest form *perceptrons*, which are connected in a network-like structure. Inputs provide sensory information, which gets evaluated in one or more layers of neurons. The resulting sum of outputs by the neuron layers generate a behavioural pattern. If certain values reach the input side, a consistent response is created as an output [18].

Reinforcement learning is probably the most complex, but also most general learning method [40]. Actions taken by the agent inevitably lead to consequences, good or bad. The evaluation of successful and unsuccessful actions is used to maximize a reward function. The agent is not being led through the learning process as in supervised learning, but instead derives the most successful actions from the rewards it gains by trying them out [41]. Therefore, reinforcement learning does not require an expert to provide the right or wrong solutions, which is an important feature of this type of learning. But more significantly, the agent is able to engage with uncertain, unknown new territory, because it learns entirely from its own experiences and not from a knowledgeable teacher. But reinforcement learning also imposes an important issue on the design of the agent architecture. Depending on the problem set and the intended use of the agent, a balance between exploration and exploitation must be maintained. The agent has to prefer actions that lead to maximum rewards in order to arrive at a certain goal. This implies that some actions which have not been tried before, might never be explored even though they could lead to even higher rewards and ultimately to the best outcome overall [43].

Agent research, and study of learning agents specifically, is a vast field of research and many concepts have been developed to improve aspects of agent architecture, communication and performance. One performance enhancement of learning algorithms, among many others, is Boosting. Boosting is a generic and often effective method of creation reliable predictions in machine learning [44]. Learning algorithms often suffer from noise in the data or small numbers of training examples [45]. Analysing and tracking the training error by use of a test set, and combining multiple resulting classifiers based on their training error score into a meta-classifier, enables the Boosting algorithm to classify instances better than individual classifiers based on noisy or small training data as shown by Schapire [46]. Human-centric evolutionary computation works with relatively small data sets, where tens or hundreds of iterations are typical, compared to non-interactive Genetic Algorithms, where thousands or millions of generations are possible depending on the computational resources. Therefore, the ability to train learning algorithms using advanced methods such as Boosting gain more importance in Interactive Evolutionary Computation as suggested by Kamalian et al. [47] in context of electronics design.

### 2.4. Collaborative multi-agent systems

Multi Agent Systems (MAS) are computational systems that integrate more than one type of agent. These agents might interact and communicate with each other and perform different or similar tasks [48]. MAS are employed when a single agent might fail to solve problems by itself, because the problem is either too difficult to encapsulate into a single agent or it is impossible to do so. For example, a human (agent) and a computational agent may interact while working on a task, or multiple different agent architectures have to be employed to solve an issue because the problem is beyond the scope of an individual agent [49].

MAS can be classified into homogeneous and heterogeneous architectures. In case of a homogeneous structure, all agents have the same underlying architecture. They only differ with regards to the environment they are in. Every agent contributes to the overall system by observing parts of the environment that other agents can not perceive. Sometimes there is an overlap between the observations made, in case those agents partially share the same part of the environment. In contrast, heterogeneous systems are made of agents of different architectures. The agents perform different tasks in different ways and complement each other. The most extreme example for a heterogeneous structure might be a MAS of human and computational agents. Homogeneous systems are relatively fast to create as they only require a single agent architecture. The advantage of heterogeneous systems is their ability to account for a wide range of different tasks, while keeping the individual agent relatively simple [49].

This research aims to combine interactive computation with autonomous computational agents. Therefore, the concept of MAS is important to understand. But as this study uses MAS in a very specific way, communication between agents and their hierarchical structure differs from common MAS and is detailed in the context of Human-Based Genetic Algorithms (HBGA).

**Human-Based Genetic Algorithms**
First introduced by Alex Kosorukoff as part of his research into knowledge management, Human-Based Genetic Algorithms [50] are an additional class of genetic algorithms. Kosorukoff describes them as a form of outsourced primary genetic operators, which are the processes of selection, crossover and mutation. Drawing the parallel to a business organization, he exemplifies outsourcing as the transfer of "ownership of a business process to an external agent" [50]. Kosorukoff further points out, that outsourcing effectively means the transfer of a function from the organization to an external agent. This function will be performed independently and unsupervised by the agent, sometimes even without any knowledge of how the agent works, which methods the agent employs and most importantly, partially or fully beyond the control of the organization. The organization only controls the choice of agents, but not their functionality. Similarly, an organizational function is introduced to coordinate the system of multiple agents.

In Human-Based Genetic Algorithms, the three primary operators are simplified into *selection* and *recombination* (merging crossover and mutation). These two main functions can be taken over by either human or computational agents – not just exclusively, but even in combination. For example, there may be a computational recombination agent, a human selection agent like in Interactive Genetic Algorithms plus an additional computational selection agent. This is the defining feature of Human-Based Genetic Algorithms. The term Human-Based Genetic Algorithms may be considered to be slightly misleading for a number of reasons. Firstly, this class of Genetic Algorithms does not exclusively incorporate human agents. Secondly, the distinguishing feature is the use of a multi-agent system not necessarily a human-based system. Kosorukoff's own publication [50] also used the term Multi-Agent Genetic Algorithms, which would be a more distinguished term and probably less prone to be misinterpreted. This is not to be confused with what Zhong et al. [51] introduced as MAGA, which is a Genetic Algorithm, where each candidate is an agent.

Kosorukoff also identified a significant implication of Human-Based Genetic Algorithms. The organizational function needs to be efficiently and carefully designed in order to allow for effective agent-agent or agent-human interaction [50]. While this indication seems to be correct, it is equally true for simple Genetic Algorithms and especially Interactive Genetic Algorithms. The program structure defines how the entities (computational or human) interact, how effectively they perform and whether it is possible to achieve any sensible, desired outcomes at all. Therefore it seems as if this seemingly generic problem is outweighed by the robustness and flexibility of a multi-agent system.

In relation to this study, it seems to be a huge advantage to be able to utilize one or more agents in addition to human selection in order to augment the before mentioned issue of user fatigue. If a computational agent performs one or many iterations of selection instead of the human user, the capability of the human is probably utilized in a better, more effective way, and therefore allows for an overall larger number of iterations, which in turn leads to a better convergence of desired and achieved results. This is one of the core ideas which this research seeks to explore.

**3. Multi-agent Human-Based GA**
This section outlines the architecture of the multi-agent human based genetic algorithm. The elements of the system are shown in Figure 1, with specific reference to the procedural city generation application outlined in Section 4.

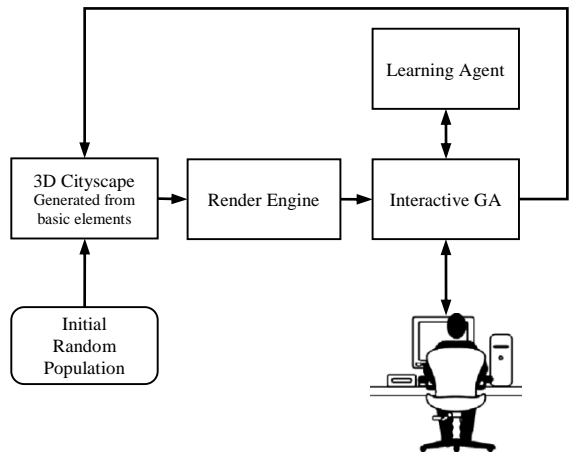

**Figure 1.** Human-Based GA architecture

An initial population of cityscapes is generated and presented to the human agent after being rendered. The human selects the candidate solutions to be used in the evolution of the next generation of cityscapes. The learning agent monitors the selections made by the human agent. The learning agent classifies new generations of candidates based on several available classification algorithms and makes new decisions following these classifications. The decisions made lead to actions, in this case selections of new candidates for future breeding. These selections are fed back into the Genetic Algorithm. While this architecture is relatively straight forward, it is still following the common view of what an agent is. Fogel [20] as well as Russel and Norvig [40], all differentiate an agent from any other software by assuming that the agent observes (at least part of) its environment, makes decisions based on these observations and takes actions accordingly. All three of those assumptions are found in the agent architecture used in this research.

The main classifiers used in this research are based on decision trees and naïve Bayes. The reasoning behind this is to evaluate different learning approaches from different classes of learning algorithms. Employing the WEKA machine learning framework allows for a fast switch between different approaches, which provides an insight into the performance of learning algorithms based on very small training sets as used in this study. Given that the aim was to counteract the effects of fatigue by running as few iterations as possible, only a small number of user selections per run were available for training the agent, before the trained classifier had to evaluate a generation of candidates itself.

This paper utilizes the C4.5 algorithm [52], which was used to induce the decision trees. Specifically, WEKA's J48 classifier, which is an open source implementation of C4.5 revision 8, created the decision tree at runtime and refined it after each selection made by the user. This is how the initially untrained agent is able observe the actions taken by the user and utilize this information to build the decision tree independent of the goal desired by the user at the start of each run. It adapts every time the whole interactive process is started and does not rely on any assumptions made by the software developer. Every time the user gives new input, the resulting generation of cities is considered a new supervised input. Subsequently, the increasing size of the training set, growing with the iterative selection process by the user, improves the classifier of the learning algorithm due to the increasing number of valid test samples, and therefore the ability of the agent to predict user preference more accurately is improved with every interactive step. This also implies that the user only gives feedback to the agent's actions by making new selections, not through a direct rewards/punishment system. Therefore, the frequency of interactive and computational runs has to be relatively high at the start. Only when the agent has received a certain number of valid test samples to build the classifier can its involvement be increased and more computational selections conducted by the agent. This ratio between interactive user selection and computational selection by the agent can be set before the start of the interactive modelling process. Nearly all experiments of this study, are based on a run of 10 interactive selections, followed by another 10 selections, where the agent and the user made selections every other time. From the $20^{th}$ selection run, the agent was responsible for 9 generations, while the user only interacted every $10^{th}$ time. While this required the user to provide an initially high number of selections, the workload was relatively quickly reduced to a very low number.

## 4. Procedural City Generation

The first important decision that led to the underlying idea of combining Interactive Evolutionary Computation and Agents for this research, was to computationally generate a model of a city. The reasoning behind this shift from manual laborious modelling to significant computational support, is the complexity and number of parameters that are necessary to create such model. While Parish and Müller [9] point out that the creation of systems of high visual complexity is an established process in computer graphics, it still requires a very high level of skill, knowledge and consumes a lot of time. Even breaking the overall model down into smaller units such as buildings, streets and layout, does not lead to a significant simplification or is less demanding towards the computer graphics expertise of the designer. Using computational approaches exclusively pose the same problem on the software developer. Many different approaches have been taken to combine smaller, simple elements into a large system. These include L-systems for plant generation [53], dress design [54], level generation for jump and run games [55] and the before mentioned L-system based *city engine* by Parish and Müller [9]. But none of these processes close the gap between highly complex systems and the inclusion of the designer in the creation process. This is where the novelty of the solution proposed in this paper lies. It enables a moderately skilled designer to create a large model of high complexity without compromising on the aesthetic demands to achieve said complexity.

In principle it is possible to use the proposed system of IGA and Agents for nearly any computer generated asset for games, film or virtual reality. A city has intentionally been selected as an example of high complexity and composed of many individual parts. The design and creation of CG cities is a complex undertaking as pointed out in previous sections.

Some of the decisions that could be considered limiting to the achievable design, such as using a square matrix to place buildings, separate land and water and drive the height and density of the city centre with its higher buildings, are indeed not limiting at all. For example, considering a map of the Manhattan peninsula of New York City and overlaying a simple square grid as shown in Figure 2 it is possible to break the complexity of Manhattan down into few small units, for example streets, buildings, water and land.

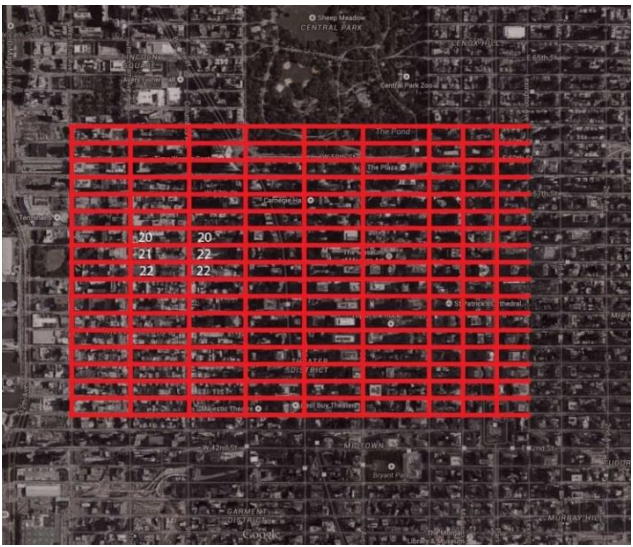

**Figure 2.** City grid overlay

Taking the idea further, the same grid overlay could be used to build a height-map of the buildings on the peninsula by assigning different height values to each grid cell as shown in Figure 3a. This is effectively reverse engineering the main spatial and aesthetic features of the city. Further, this approach does not limit the street layout to be square and grid like. Figure 3b is based on an underlying square grid, but shows typical non-square, European like cities, which evolved from a city centre and spread outwards like a web. And this reverse approach is the foundation of the city model as discussed in this study. A simple grid structure is introduced to represent buildings, land and water. A street-map is used to build the network of streets between the buildings on land, and finally an occupancy grid is used to indicate the presence of water.

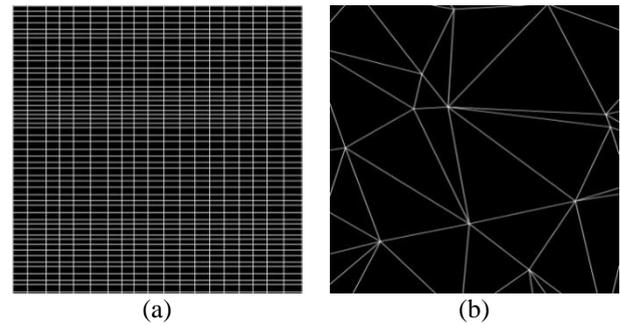

(a)             (b)

**Figure 3.** Street layout styles

A height-map is applied to the buildings in the grid, which drives the height of the buildings plus a pre-defined variance so that the elevation does not appear too uniform, but believable and aesthetically similar to what is expected based on looking at actual cities. Further, only simple instances of buildings are used during the design process. This is for two reasons, one being the very practical realization that creating and rendering times have to be as short as possible to reduce additional user fatigue. Secondly, the instances can easily be replaced with prebuilt or purchased models of very different architectural styles. The applicability of this approach is therefore much broader than the presented prototype might suggest. Further investigation is of course needed to verify this, but again the limiting factor is predominantly the scope of this study and not the presented process itself.

The GUI was written in Java using the Processing framework to provide some graphics functions. There are only two input parameters, which reflect the two main parents of each population of the Genetic Algorithm. The GUI therefore simply presents a number of candidate solutions from which the user selects the two chosen parents. The GUI is therefore represented by the figures presented in the results section of this paper. The motivation behind this simple mode of interaction is the desire to reduce user fatigue. Additional inputs would require additional user attention. And as the models, which the user designs, grow in complexity, the risk of extending the required attention and interaction beyond the point of human capacity is rising as well. Minimizing the number of required interactions per iteration seemed the most appropriate way to focus the results of this study on said fatigue and avoiding it by adding agents.

The resulting city model is being kept fairly simple in order to make rendering of 9 candidates for a number of iterations feasible. This is just a limitation of the available hardware and could easily be changed in a commercial environment by using a small render farm. But to manage the rendering times (which are purely a result of the commercial render engine and not the software system presented here), only simple geometric primitives such as cylinders and cubes have been used for the object instances. Also, the texture is very small and simple, due to the limited memory available in the test system. Given that the software system has been designed to use instances of small building blocks instead of copies of existing models, the process of changing from this simple type of geometry to a fully textured, shaded, high polygon model for each house is just the mere adjustment of one configuration parameter. The system to produce feature film quality renders has been established and poses no limitation to the validity of the results. It is just a matter of using high end workstations, which could also be done in future research, assuming that the resources would be accessible.

A typical model, as it is shown to the user for selection, is shown in Figure 4. Some of the defining features include a New York style street system which divides the city into square blocks, some water visible in the far background as well as a distinct city centre on the far left with a number of larger than average buildings surrounding it.

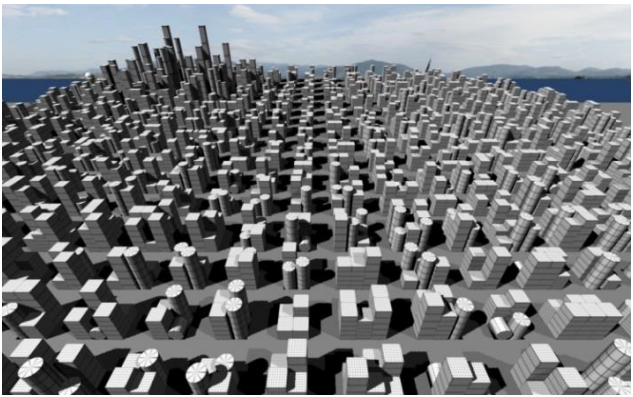

**Figure 4.** Typical candidate solution

The DNA string used in the implementation of the Genetic Algorithm is made of the following parameters:

- Ground: 2D integer array, reflecting land or water
- Heightmap: 2D integer array, height of buildings
- Streets: 2D integer array, street map layout
- Buildings: 2D integer array, type of building
- City: 2D integer array, location of city center(s), a few distinct buildings

One of the necessities for inducing decision trees is to keep the number of attributes as low as possible, while still arriving at a solution in form of a usable classifier [56]. To accommodate this requirement, the parameters for the agent are abstracted from the candidates DNA.

Instead of using each attribute of the DNA, some have been consolidated. For example, the height map is a grid-like structure represented as a 2D array in the DNA, which has been pre-processed into three parameters, namely average building height, number of buildings and height of buildings in the city centre (or scale of the buildings compared to the average height in the remainder of the city). This reduces the number of parameters from originally 10,000 cells in the 2D grid, holding the height value for each individual cell, to just 3 attributes, which make induction of a decision possible for the agent.

In summary, the attributes for the agent are:

- Land/Water ratio
- Street type (European or New York style)
- Average building height
- Number of buildings
- Height of city centre

Additionally, the agent receives the class value, which indicates whether this instance of the training set has been selected or rejected. Finding the correct class value in new instances is ultimately the task of the agent.

## 5. Experimental design

The overall experimental process for procedural cities involves three different runs, namely Random, IGA and HBGA, but not necessarily in this order so that the user does not attempt the design with a preoccupied idea of how the software might perform. For all three variants, the user interaction is exactly the same, so that the underlying strategy is hidden from the user. The differences between the three variants are detailed in the following subsections.

In any case, each iteration of the user interaction starts with a set of 9 rendered city models. These are created by the software system depending on the underlying model, for example random selection, interactive genetic algorithm without agent and interactive genetic algorithm with agent. The user selects two parents with the mouse (both show a coloured border around the selection to give visual feedback) and starts the algorithm with a press of the *space* button. Everything else such as log information, render file creation and sub-processes for the render engine are hidden from the user. This keeps the user interface as simple as possible, with the idea to utilize maximum user focus for the actual task of selection.

After computation, which includes creating the new population of 9 cities, writing the render archives and rendering the resulting images, the user is presented with the next generation ready for selection. Each generation typically takes around 10 seconds to render the cities and the average time taken for a user to conduct an evaluation was an additional 5 seconds.

Each individual evaluation was conducted based on a predefined set of goals. Some of the requests that where made include relatively specific elements such as:

- A city with a lot of blue water, a city centre with very high buildings on the left and some smaller buildings on the right hand side of frame
- A city without water, multiple city centres with few or no small buildings visible in frame
- A city without water, no distinct city centre and only very small buildings

Other requests were kept abstract using high level description of the desired features. Again, this was designed so that the difference between a tight goal driven, perhaps client based approach and in difference a free creative design, independent of any strong prerequisites and maybe just loosely based on a design idea could be simulated. The latter examples include:

- A harbour city with a large population
- A city in a valley with a suburban feel
- A city by the sea with a lot of tourism

A number of results were recorded for each different approach. These findings include the run times for the overall process from start to finding a result that the designer deemed final, the number of iterations required to get to the final result, and also the subjective feeling after performing a full run. While the latter is not necessarily representative for the quality of the algorithm, with regards to user fatigue, it might provide an idea whether the software system is successfully reducing the workload on the user.

## 6. Results

The experimental process involved three different runs, namely random selection, Interactive Genetic Algorithm without agent and Human-Based Genetic Algorithm including a computational agent.

### 6.1. Random selection

Initial testing did not show any promising candidates, even after a larger number of runs. In a few rare cases though, a random sample early on in the process could have been accepted under the assumption that the brief was not taken too rigorously, but none of the candidates resembled the previously stated goal for that run exactly. But the waiting times for each generation are relatively high, with only a slim chance of randomly striking an acceptable solution. Involving the human into the full process, just to create a random control was deemed to be impractical. Instead, batch processing was realized and the user went through the results only. While this approach did not measure the fatigue generated by the actual (redundant) selection plus render times for every generation, having to go through 900 pictures proved to be very tiring and frustrating.

Two runs were conducted, and the experiment was not continued for the initially envisioned 10 runs. The time it took to evaluate the images seemed to justify the conclusion that random selection does not necessarily lead to a result within a practical time frame. While it could be argued that manual creation takes much longer, this study's focus is on computational solutions, and the run times in the following sections demonstrate the difference between a manual random and a computational approach.

One interesting finding is the relevance of the design goal though. When looking for a very specific outcome, for example a large city by the water with many high rise buildings on the left of screen and some flat buildings on the right, it was clear after 100 generations (or 900 images) that no solution had been found. But when the goal was set in a more abstract way, without any specific requirements, for instance large harbour city, some of the candidates seemed to fulfil that brief at least very loosely. But in saying this, it seems necessary that the set goal has to be specific enough, so that any broad interpretation, which might include many very different solutions, is avoided. A loose goal setting would not provide any contestable results with regards to the specific research question of this study. This research is concerned with the difference between pure Interactive Evolutionary Computation and Human-Based Genetic Algorithms to address fatigue.

### 6.2. Interactive GA

This section outlines a typical run using Interactive Genetic Algorithms without the support of a computational agent. This section outlines an example case of the system in use. The predetermined goal was very specific, stating that a city with a lot of blue water, a distinct city centre on the left and much smaller buildings on the right was to be created. Figure 5 is the first generation of the example generated from the random seed.

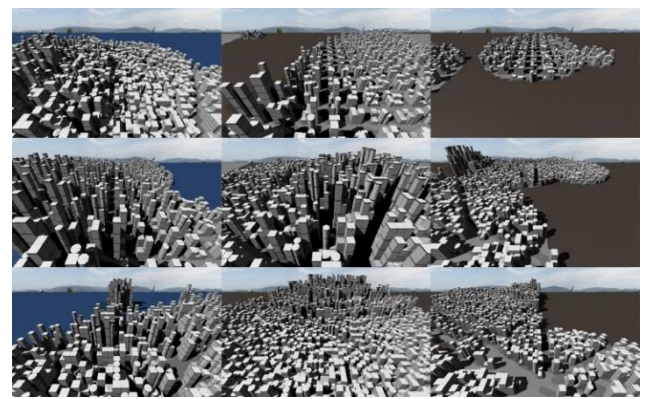

**Figure 5.** Generation 1 of an IGA run

Typically, the first generation contains a number of mixed results, here with 6 of 9 candidates containing just land and no water. Consequently, the user selected candidates top/left and bottom/left.

As a result of mainly choosing candidates that had water present, the user improved the number of available options in subsequent generations. Figure 6 shows generation 10, which has significantly more cities with water visible.

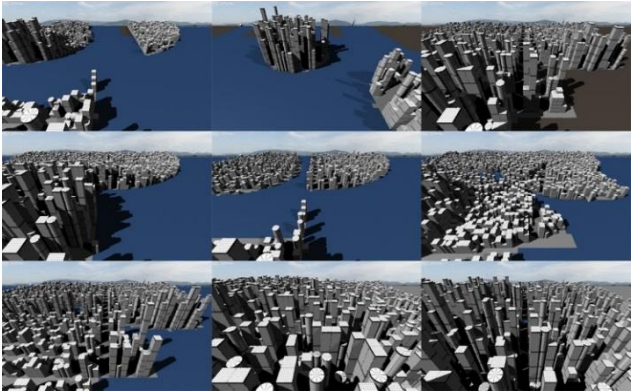

**Figure 6.** Generation 10 of an IGA run

Another 10 generations later, a number of candidates with water appear and high buildings are emerging on the right hand side, and on the left and somewhat smaller buildings on the right hand side. This is shown in Figure 7.

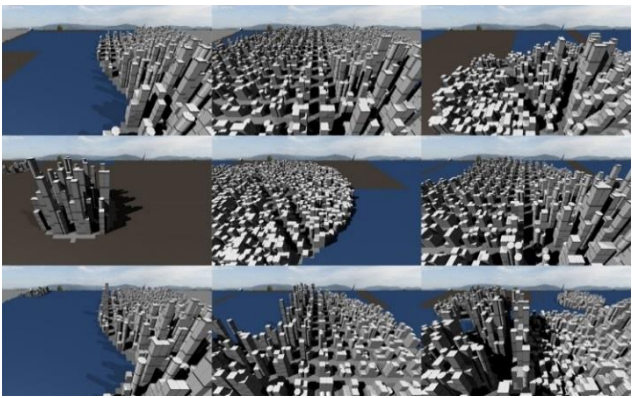

**Figure 7.** Generation 20 of an IGA run

Figure 8 illustrates generation 30 of this exemplary IGA run, a few issues can be observed, which are based on the slow convergence of the Genetic Algorithm. In this case, 8 of the 9 candidates contain water. One candidate has high buildings on the left side, and the majority of candidates had the buildings on the left. But none of the candidates presented the required properties as outlined before the run started. Accordingly, another 10 generations were required in order to achieve the first promising results, as depicted in Figure 9.

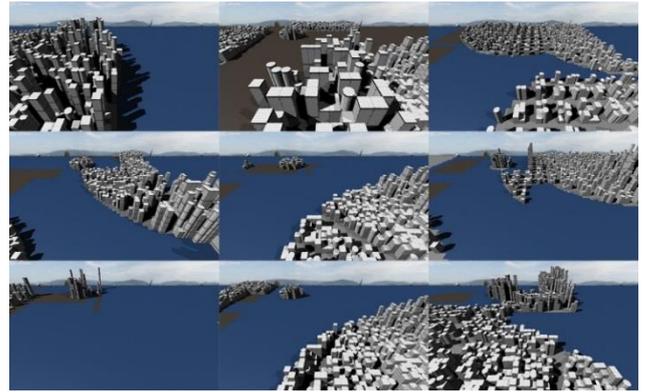

**Figure 8.** Generation 30 of an IGA run

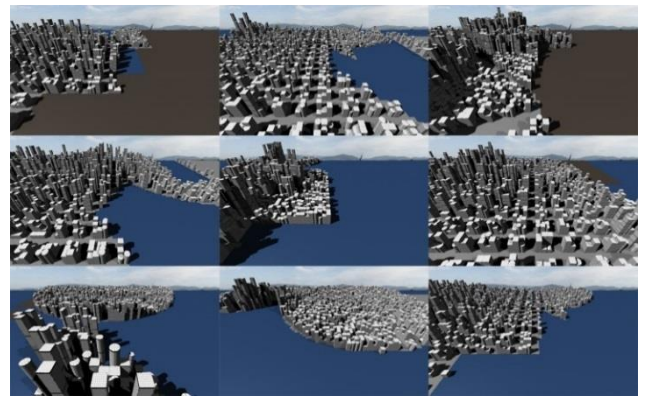

**Figure 9.** Generation 40 of an IGA run

After 7 additional generations, the final candidate was found. Figure 10 shows the originally requested water, a distinct centre with high buildings on the left and low-rise buildings on the right side of frame. While the result satisfied all criteria of the brief, it required 47 generations to achieve it.

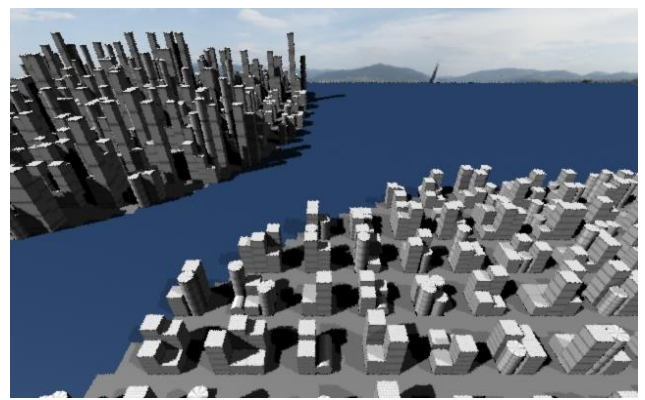

**Figure 10.** Selected final solution

Overall, the Interactive Genetic Algorithm without an agent was tested in 36 runs with an average of 37 generations per run and the average time per run was 18 minutes. The standard deviation was 11 generations, which shows a fairly wide spread. The smallest run was only 1 generation, although this was deemed a special case and likely due to the subjective nature of aesthetics combined with an element of 'luck' arising from the stochastic nature of the algorithms. After finding a suitable candidate right in the first generation, a few more iterations were conducted, which showed even more promising results. The fact that the initial conclusion in generation 1, was later reviewed and seemingly better solutions were found in subsequent generations, underlines the implications of judging aesthetics based on a high level project brief (which was given for this particular run). A more detailed brief would probably have led to more iterations in the first place.

### 6.3. Human-Based GA

The Human-Based Genetic Algorithm was tested in 36 runs, similar to the Interactive Genetic Algorithm and with the same predetermined goals, as discussed in the previous section. The average number of generations was 52 with a standard deviation was 14. The maximum was 91 generations and the minimum 21. The average time per run was 12 minutes. It is noteworthy that the total number of iterations is higher compared to IGA, but this higher number includes both interactive and computational generations conducted by the human designer and the autonomous agent respectively. The mean number of generations using HBGA that required user interaction is 18 and as such is about half of the number required when using the IGA. Table 1 shows an overview of the statistics of all runs conducted using the Human-Based Genetic Algorithm compared to the Interactive Genetic Algorithm. For each run, the same brief was given based on examples shown in section 5.

Table 1. Comparison IGA vs. HBGA

| Approach | Performance | | | |
|---|---|---|---|---|
| | # runs | # generations (interactive) | # generations (total) | time (average) |
| IGA | 36 | 37 | 37 | 18 |
| HBGA | 36 | 18 | 52 | 12 |

The higher number of generations in total indicates that a greater number of possible solutions are being explored, though a smaller number are being evaluated by the user in person.

Different initial mutation rates for the Genetic Algorithm were verified, with the majority of runs conducted at 0.2 probability. This showed a good performance in terms of relatively quick convergence, without the issue of getting stuck in local maxima. The latter was experienced at initial mutation rates of 0.02. At this low rate, the system seemed to produce little diversity even after only a few runs and the user could not achieve the predetermined goal as most candidates looked very similar and left no room for additional evolutionary breeding. Such lack of diversity is perhaps to be expected given the relatively small population size.

## 7. Discussion

It is interesting to see that the average number of generations using the Human-Based Genetic Algorithm is nearly one and a half times of the number of generation conducted using the Interactive Genetic Algorithm. This is not unexpected though. In case of the Interactive Genetic Algorithm, the user has to run every generation interactively. The time consumed per iteration is about 10 seconds render time plus user decision time, which was typically about 15 seconds. This means, a run took on average just over 18 minutes. Comparing this to the Human-Based Genetic Algorithm, the time for the first 10 runs is identical. But after that, the non-interactive generations, driven by the computational agent take virtually no time (under 1 second) for the decision making process and only the last generation that is to be presented to the user for interactive selection again, needs to be rendered, which takes the aforementioned 15 seconds. Therefore, many additional generations can be run in the same time, which the user seems to take advantage of in case of the Human-Based Genetic Algorithm.

For the overall process it can be said, that if the user is less pleased with the results returned by the agent, the user will select candidates that are different, rather than similar. This triggers the agent to change course as well, running a lower level of confidence due to the inherent inability to predict the sudden random selection by the user, which in turn creates more diversity through increased mutation probability. Consequently, this allows the designer to choose a more intuitive, even unstructured approach to the modelling process, and a carefully, clearly planned execution is not a requirement anymore. The designer is essentially able to use playful discovery without endangering the end product. In a random or manual approach, this would cost either a lot of time, as many hundreds or thousands of parameters would have to be adjusted, or it would be impossible, given that a certain appearance of the buildings in the skyline can only be altered by changing the layout of the city blocks and the street pattern.

A few interesting cases could be observed, where an originally weak and seemingly unstructured response from the algorithm in the first 10 iterations is altered by the support of the software agent within a few iterations. The software agent suddenly drives the designs into a different direction from what the Genetic Algorithm did, and closely follows what it identified based on the human selections. Therefore, the overall system performed better than its individual components and provided a better user experience. For example, the brief was to create a city without water and flat buildings with no distinct city centre. The Genetic Algorithm showed a high number of candidates with water, an average of 7 out of 9 per generation. The user inevitably selected candidates with land and no water. While the Genetic Algorithm continued to present candidates with water in higher proportion as shown in

Figure 11, the agent's classifier was trained by the user selection.

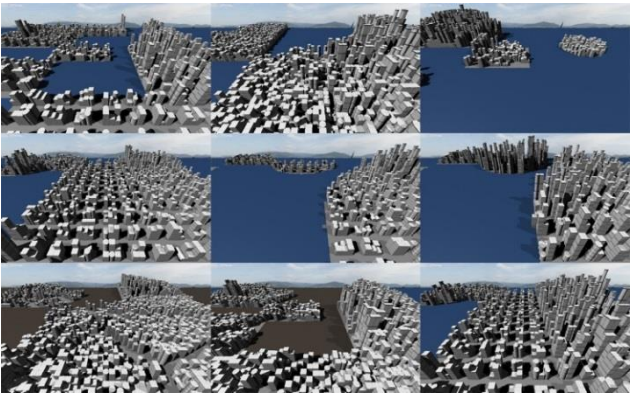

**Figure 11.** Generation 10 of HBGA run presenting mostly water, contrary to the brief given

Once the agent came into effect, after only four additional generations most of the candidates were containing land. Even the candidate solutions still containing water had proportionally more land and buildings visible, as shown in Figure 12.

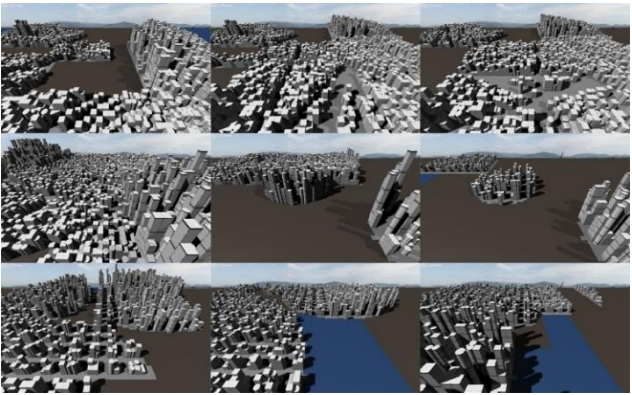

**Figure 12.** Generation 14 of the same HBGA run as above, showing mostly land-based candidates

It was a bit surprising that the user did not always follow a straight approach towards the goal. For example, if a harbour city model was requested, quite a few selections involved no water at all. There are a few possible reasons for that. First, perhaps most of the other parameters did not fit the brief from the perspective of the user. Or, the fact that relatively little effort to create a new city layout was required, compared to manual modelling, led to a more playful attitude. Overall, it seems that the user is more adventurous using the Human-Based Genetic Algorithm, changing direction a few times, for example from water on the left to water on the right when asked to make a harbour city. One would probably not attempt a drastic change after many man hours of modelling manually, as a larger diversion from the original layout might require a re-start of the whole manual modelling process. Due to the support of the agent and the relatively fast 'modelling' approach, there seems to be a lower boundary for otherwise significant changes. It seems that interactivity on one hand, but also the agent reducing fatigue on the other hand, allow for more user iterations and therefore exploration of different solutions.

The core of the agent architecture used in this study is the decision tree, which is induced at run-time and refined in subsequent iterations by use of the additional selections made by the user. Figure 13 shows the root node of the decision tree after 14 user-driven iterations.

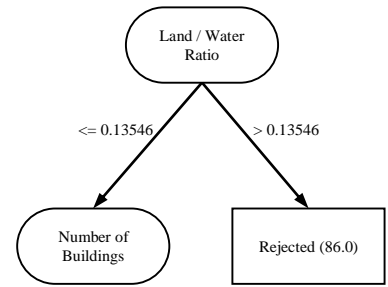

**Figure 13.** Decision tree visualization

The root attribute is land/water ratio, which gives a clear indication that any candidate with more water than 14% is to be rejected, which is true for 86 of all instances of the training set. The goal of this run was in fact to find a city with no water, and based on the training set, the classifier preferred any candidate with less than 14% of water. The full tree has 7 branches and shows a high rate of confidence (88% correctly classified instances), although most of the candidates were rejected right at the root, based on the amount of water compared to land hence only the root node of the tree is shown. This is one indication of a useable classification tree, however, some of the other decision trees that have been examined throughout this study, were not as clear and had a lower level of confidence. There has been no clear indication that the J48 classifier produces viable results in every case. Sometimes the tree is not able to reliably identify candidates with a high confidence and the score was only around 60%. However, this is still a marginal improvement over a 50/50 'coin toss', so there is some value in the use of the agent. It seems that perhaps the noise of the training data due to the low number of instance compared to other data mining tasks, might be a contributing factor. Hall et al. [56] discuss this as a possible issue, and other work using J48 has identified that when the number of instances is low compared to the number of attributes, the J48 classifier becomes of limited use [57], though one possible solution to this problem is the use of synthetic data [58]. In the instance of the procedural city generation, the classification is conducted on 5 attributes and typically the training set would include 10 initial instances. The selection of J48 was initially based on its popularity as a classifier, however a potential solution to improve the effectiveness of the classification might be the introduction of an alternative classifier, which could be part of future research and the matter requires further investigation. The WEKA framework allows for relatively easy adaption of different algorithms.

## 8. Conclusion

This paper has presented initial results comparing the use of a simple Interactive Genetic Algorithm with a multi-agent implementation of a Human-Based Genetic Algorithm with the goal of determining whether the computation learning agent in the HBGA has the potential to reduce user fatigue. Results indicate that the use of the HBGA allows greater exploration of the design space in a shorter space of time. This suggests that there may be lower overhead placed upon the human user which and that there is potentially less fatigue experienced to achieve the same goal. However, further work with more subjects and quantification of the fatigue

In terms of actual fatigue, an interesting observation is the influence of soft factors such as positive emotions. Fatigue seems not just to be based on attention span and focus, but also to be compensated by subjective positive emotions. Evaluating the candidate solutions presented by the computational agent seems to positively engage the designer more than when evaluating those presented by the IGA. The process of using interactive evolutionary concepts, which allow the user to observe convergence towards selected goals with each iteration, could be almost described as playful.

Looking at the number of iterations run by the Human-Based Genetic Algorithm compared to pure Interactive Genetic Algorithm, it seems that the user might be happy to allow more iterations, if they are not interactive but run by an agent. It seems not to be about keeping the maximum number of iterations low, but more about optimizing the final result within a certain time frame. The average time of the runs between IGA and HBGA were very similar, which could indicate that the user is more driven by time consumed, rather than the number of selections that have to be made by either the human or the agent. This might hint that the driving factor is indeed fatigue or attention span, and that a computational agent helps to optimize the result by running additional iterations. Based on this prototype, it looks as if the system of Interactive Evolutionary Computation and agents shows some promising benefits towards goal optimization, and a wider study could probably confirm this indication and provide additional insights.

The results of this study do not show conclusive evidence that agents lead to consistent improvements of Interactive Genetic Algorithms. But there are some indications that this approach has advantages. First, there are some promising signs when adding agents to the interactive process, for example the cases where the Genetic Algorithm seemed to suffer from a high mutation probability, which lead to a high diversity and no clear convergence. Once the agent ran some of the generations, a clear shift in direction towards the previous user selection was observable. This needs further proof, which a quantitative experiment could provide. Second, the observation that the user seemed to enjoy the interactive process more, once the agent was engaged, could prove to be a valuable insight. While this needs further investigation as well, looking at the psychological aspects of perceived intelligence by a computational system could provide additional value. This seems like the next logical step in understanding the user experience of Interactive Evolutionary Computation better.